%% file: main.tex
\documentclass{article}
\usepackage[T1]{fontenc}
\usepackage{iclr2027_conference,times}
\usepackage{amsmath,amssymb}
\usepackage{booktabs,tabularx,array,multirow}
\usepackage{graphicx}
\usepackage{microtype}
\usepackage{url}
\usepackage{algorithm}
\usepackage[noend]{algpseudocode}
\usepackage{xcolor}
\usepackage{enumitem}
\usepackage{xspace}
\usepackage[hidelinks]{hyperref}
\usepackage{float}

\newcommand{\SSR}{\textsc{SSR}\xspace}
\newcommand{\NMSE}{\textsc{NMSE}\xspace}
\newcommand{\LLaDA}{LLaDA\xspace}
\newcommand{\Dream}{Dream\xspace}
\newcommand{\Udense}{U_{\mathrm{dense}}}
\newcommand{\Aseq}{A_{\mathrm{seq}}}

\newcommand{\pp}{\ensuremath{\,\mathrm{pp}}\xspace}
\newcolumntype{Y}{>{\raggedright\arraybackslash}X}
\newcolumntype{C}[1]{>{\centering\arraybackslash}p{#1}}
\setlist[itemize]{leftmargin=1.3em,itemsep=1pt,topsep=2pt}

\title{What Does FFN Compression Change\\
Downstream? Same-State Causal Restoration\\
in Diffusion Language Models}
\author{Shaurya Omar\\Indian Institute of Technology, Roorkee}

\iclrfinalcopy 
\begin{document}
\raggedbottom
\maketitle
\lhead{Under review as a conference paper at ICLR 2027}

\makeatletter
\renewenvironment{abstract}{%
  \vskip.075in\centerline{\large\sc Abstract}%
  \vspace{0.5ex}\vspace{1.65pt}%
  \begin{list}{}{\leftmargin41.8pt\rightmargin39pt}\item\relax\setlength{\baselineskip}{10.963pt}%
}{\end{list}\vskip 1ex\vspace{16.67pt}}
\makeatother

\begin{abstract}
Diffusion language models (DLMs) enable flexible, parallel generation, but their iterative denoising remains computationally expensive, motivating increasingly aggressive compression. Existing compression objectives largely measure how well compressed computation approximates the original locally, but local error does not reveal which removed computations actually matter to the downstream denoising trajectory. We introduce Same-State Causal Restoration (\SSR), which restores the original FFN on the exact current input reached by the compressed model and measures how the resulting trajectory changes. To our knowledge, this is the first direct measurement of the same-current-input closed-loop effect of removed FFN computation in DLM compression. Across \LLaDA-8B-Instruct and \Dream-v0-Instruct-7B, compressed-side state ranks this downstream effect substantially better than local \NMSE at fixed denoising phase, while controlled interventions show that correction structure matters beyond magnitude. Using task-label-free calibration, \SSR freezes a single restoration window for held-out inference. Under aggressive \LLaDA compression, restoring only four transi-\linebreak tions recovers 89.9\% of the lost accuracy while retaining an estimated 36.8\% whole-model MAC saving and outperforming an equal-budget local-error base-\linebreak line. \Dream further shows that restoring dense behavior and repairing the final task are distinct outcomes.
\end{abstract}

\section{Introduction}
FFN compression inside an iterative DLM can change both the current computation and the states on which later denoising transitions operate. Existing DLM efficiency methods exploit timestep- and state-dependent structure, while causal-tracing work intervenes on DLM states or components \citep{zhang2026quantdllm,oba2026stopping,park2026knowledge,yao2026timerome}. Neither directly measures the compression-specific downstream effect of reinstating the exact FFN computation removed by compression on the trajectory the compressed model actually visits.

Once dense and compressed histories diverge, ordinary FFN comparisons change both operator and input. \SSR instead evaluates the original and compressed FFNs on the same current restoration-branch input and follows the resulting continuation (Section~\ref{sec:method}), isolating operator substitution from trajectory-induced input shift.

The distinction is consequential: under aggressive \LLaDA compression, a single globally frozen \SSR window recovers 89.9\% of the dense--compressed accuracy gap while retaining an estimated 36.8\% whole-model MAC saving, and exceeds the equal-horizon local-\NMSE/final-four schedule by 5.05\pp.

We organize the paper around three quantities that are easy to conflate:
\begin{center}
\small
\textbf{local error} \ $\neq$ \ \textbf{downstream effect of missing computation} \ $\neq$ \ \textbf{task repair}.
\end{center}

\begin{figure}[H]
    \centering
    \vspace{-4.9pt}\includegraphics[width=\linewidth]{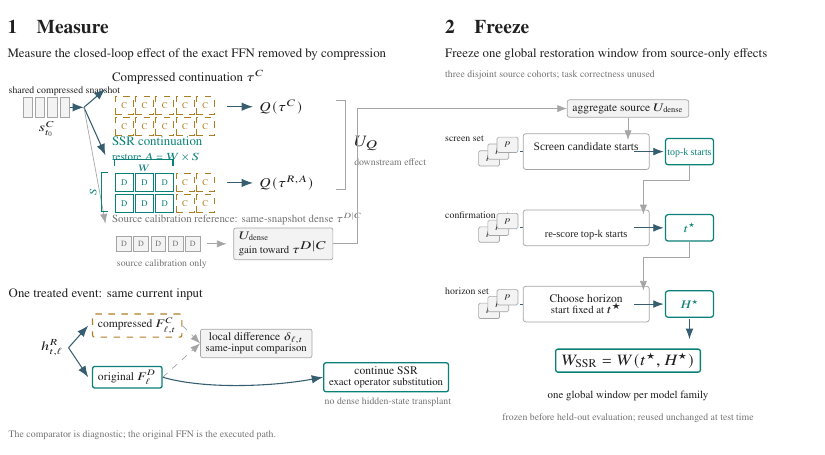}
    \caption{\textbf{Same-State Causal Restoration: measure and freeze.} From one compressed snapshot, exact \SSR substitutes the original FFN on the same current branch input and follows the resulting continuation; the local FFN difference is only a local-error signal. Task-label-free source $\Udense$ effects select one global window, frozen before held-out evaluation. No dense hidden state is copied.}
    \label{fig:overview}
\end{figure}
\vspace{0.94pt}

Local error measures present mismatch; downstream effect measures the later change after restoring missing computation; task repair asks whether that future crosses the final correctness or validity boundary.

\paragraph{Measure.}
\SSR defines a compression-specific downstream-effect estimand by restoring the compressor-removed original FFN on the current endogenous branch state and following the resulting future.

\paragraph{Explain.}
At fixed phase, compressed-side state ranks measured effect better than local \NMSE, while norm-matched $\delta$ interventions test correction structure beyond magnitude (Figure~\ref{fig:mechanism}).

\paragraph{Freeze and transfer.}
Task-label-free source effects select one global window per family, frozen before held-out evaluation. Aggressive \LLaDA shows strong task repair, moderate \LLaDA supplies a tie boundary, and \Dream improves full-dense fidelity without a conclusive task effect.

\paragraph{Contributions.}
\begin{itemize}
\item \textbf{A compression-specific downstream-effect estimand.} \SSR controls the operator/input confound by evaluating original and compressed FFNs on the same current branch input and measuring the resulting closed-loop effect, without transplanting a dense hidden state.
\item \textbf{Cross-family evidence against magnitude-only explanations.} At fixed phase, compressed-side state ranks measured downstream effect better than local \NMSE; norm-matched interventions show that correction structure matters beyond magnitude.
\item \textbf{Frozen held-out restoration under retained compute.} Task-label-free calibration freezes one global window; aggressive \LLaDA recovers 89.9\% of the accuracy gap while retaining an estimated 36.8\% whole-model MAC saving, with c14 and \Dream defining complementary boundaries.
\end{itemize}

\section{Related Work and Novelty Boundary}
\paragraph{DLM compression and adaptive computation.}
DLM-native efficiency work already exploits denoising-specific structure. Quant-dLLM calibrates to timestep-dependent masking; DLLMQuant reports quantization-error accumulation across iterations; and Quantization Meets dLLMs systematically characterizes post-training quantization behavior \citep{zhang2026quantdllm,xu2026dllmquant,lin2026quantization}. Adaptive methods cache, skip, or stop computation from token or trajectory state \citep{jiang2026d2cache,zhu2026esdllm,li2026adadlm,oba2026stopping}. Layer-collapse and activation-sparsity studies further show non-uniform DLM/FFN structure \citep{conzelmann2026layercollapse,szatkowski2026activation}. These works optimize or characterize efficient execution; they do not measure the downstream effect of reinstating the exact computation removed by compression.

\paragraph{DLM causal interventions.}
Causal tracing has also reached DLMs. \citet{park2026knowledge} restore clean-run hidden states or MLP contributions for factual localization, while TimeROME-DLM traces and edits temporally localized DLM states \citep{yao2026timerome}. These works establish DLM causal intervention as prior art, but intervene on transplanted mediators or editing coordinates rather than the compressor-removed original FFN evaluated on the current compressed/restored branch.

\paragraph{Compression causality and exact gap.}
In autoregressive models, When Reasoning Meets Compression uses attribution patching to localize compression-sensitive components \citep{zhang2026reasoningcompression}, while \citet{gu2026routeflip} causally decomposes quantization damage. Thus causal analysis of compression is not itself new. The remaining DLM-specific question is whether, at a state actually reached by a compressed/restored trajectory, reinstating the exact FFN computation removed by compression changes the future continuation. To our knowledge, prior DLM work has not jointly measured this same-current-input closed-loop effect of the exact compressor-removed FFN and used task-label-free source effects to freeze a selective restoration window for held-out execution.

\begin{table}[H]
\caption{\textbf{Closest conceptual neighbors and the intervention boundary relative to \SSR.} Existing work covers DLM compression, adaptive computation, or causal intervention; \SSR targets the compression-specific same-current-input continuation effect.}
\label{tab:neighbors}
\centering
\small
\renewcommand{\arraystretch}{0.94}
\begin{tabularx}{\linewidth}{@{}p{0.22\linewidth}p{0.31\linewidth}Y@{}}
\toprule
\textbf{Work} & \textbf{Object / signal / intervention} & \textbf{Boundary relative to \SSR} \\
\midrule
Quant-dLLM; DLLMQuant; Lin et al. & DLM compression / sensitivity-aware PTQ & optimizes or characterizes compression; no exact removed-FFN continuation effect \\
SureLock & converged-token compute / local posterior stability & skips query + FFN; no compressed-vs-original FFN contrast or exact restoration \\
Layer Collapse; Szatkowski et al. & FFN/layer compressibility, redundancy, activation sparsity & structural diagnostics; not causal downstream-effect measurement \\
Park \& Jo & DLM factual localization / clean-run state or MLP restoration & transplanted mediator; not compressor-removed operator on current branch \\
TimeROME-DLM & temporal DLM causal tracing / editing & causal DLM state intervention; not compression-specific operator substitution \\
\SSR & exact original FFN on current compressed/restored branch & same-current-input signed effect + frozen task-label-free transfer \\
\bottomrule
\end{tabularx}
\end{table}
\vspace{-9.67pt}

\section{Same-State Causal FFN Restoration}
\label{sec:method}
\subsection{Measure: downstream effect on the compressed trajectory}

\paragraph{Fixed compressed model.}
Write one deterministic denoising transition as
\begin{equation}
s_{t+1}=\Phi_t\!\left(s_t;F_{1,t},\ldots,F_{L,t}\right),
\label{eq:transition}
\end{equation}
where $s_t$ contains the partially denoised sequence and every sampler or persistent compressor variable needed to reproduce continuation from transition $t$. The original FFN at layer $\ell$ is $F^D_\ell$; a fixed compressor replaces a predetermined layer set $S$ by cheaper operators $F^C_{\ell,t}$. For convenience set $F^C_{\ell,t}\equiv F^D_\ell$ for $\ell\notin S$, so $F^C$ denotes the complete baseline FFN family. All non-FFN computation is unchanged; persistent compressor-state arguments are suppressed. Original and compressed FFNs share the residual-stream interface; representation-changing compressors require an alignment map and are outside the present estimand.

\vspace{0.69pt}\paragraph{Local error.}
On current FFN input $h$, the missing same-input correction is
\begin{equation}
\delta_{\ell,t}(h)=F^D_\ell(h)-F^C_{\ell,t}(h).
\label{eq:delta}
\end{equation}
The reconstruction baseline is measured on the ordinary compressed branch. For prompt $p$ and transition $t$, with evaluated token inputs $H^C_{p,t,\ell}$,
\begin{equation}
E_{p,t}(S)=
\frac{\sum_{\ell\in S}\|F^D_\ell(H^C_{p,t,\ell})-F^C_{\ell,t}(H^C_{p,t,\ell})\|_F^2}
{\max\!\left(\sum_{\ell\in S}\|F^D_\ell(H^C_{p,t,\ell})\|_F^2,\varepsilon\right)}.
\label{eq:nmse}
\end{equation}
We call $E$ \emph{local \NMSE}, with $\varepsilon=10^{-12}$ as executed. It measures mismatch now and contains no continuation. Local \NMSE is defined on the pre-restoration compressed-branch state and used only as an offline local-error baseline for window selection; computing it requires the original FFN, so it is not a zero-overhead online routing signal.

\vspace{-8.84pt}\paragraph{Why ordinary dense--compressed differences are confounded.}
Once dense and compressed histories diverge, a dense--compressed FFN difference changes both the operator and its input. Add and subtract $F^D_\ell(h^C)$:
\begin{equation}
\underbrace{F^D_\ell(h^D)-F^C_{\ell,t}(h^C)}_{\text{ordinary dense--compressed difference}}
=
\underbrace{F^D_\ell(h^C)-F^C_{\ell,t}(h^C)}_{\text{same-input operator difference}}
+
\underbrace{F^D_\ell(h^D)-F^D_\ell(h^C)}_{\text{input-shift term}}.
\label{eq:decomp}
\end{equation}
Evaluating both FFNs on one shared current input removes the input-shift term exactly. This isolation is local to the treated event: after substitution, later states are allowed to diverge, and measuring that divergence is the point of \SSR.

\vspace{-4.54pt}\paragraph{SSR continuation.}
Let $W$ be a restoration window, $A=W\times S$, and $t_0$ its first transition. We save the complete compressed snapshot $s^C_{p,t_0}$ and start both continuations from that snapshot. The restoration branch uses
\begin{equation}
F^{R,A}_{\ell,t}(h)=
\begin{cases}
F^D_\ell(h), & (t,\ell)\in A,\\
F^C_{\ell,t}(h), & \text{otherwise.}
\end{cases}
\label{eq:restoreop}
\end{equation}
Let $\tau^C_p$ and $\tau^{R,A}_p$ be the complete compressed and restored continuations after the shared snapshot. For any endpoint $Q$ oriented so that larger is more of the target property,
\begin{equation}
U_Q(p,A)=Q(\tau^{R,A}_p)-Q(\tau^C_p).
\label{eq:utility}
\end{equation}

\vspace{-5.49pt}
At every treated event, $F^C_{\ell,t}(h^R)\rightarrow F^D_\ell(h^R)$ is evaluated on the identical current restoration-branch input and compressor state. ``Same-state'' is therefore event-level, not a claim that the whole trajectory stays fixed. Earlier restorations may change later states, so a window is a joint intervention rather than a sum of isolated one-event effects. Original-FFN restoration is locally faithful by construction; its signed closed-loop effect may still be positive, negative, or zero. ``Causal'' refers only to this controlled operator substitution and paired continuation contrast; the state observables analyzed below are not themselves intervened on.

\vspace{0.64pt}
\paragraph{Same-snapshot dense reference.}
During task-label-free source calibration we also compute $\tau^{D\mid C}$, the dense continuation from the same compressed snapshot. Let $Y(\tau)$ denote a continuation's final generated sequence. For the $m$ generated positions, define $\Aseq(Y,Y')=m^{-1}\sum_{j=1}^m \mathbf{1}[Y_j=Y'_j]$. The calibration endpoint is
\begin{equation}
\Udense=
\Aseq(Y(\tau^{R,A}),Y(\tau^{D\mid C}))
-\Aseq(Y(\tau^C),Y(\tau^{D\mid C})).
\label{eq:udense}
\end{equation}
$\tau^{D\mid C}$ is a functional reference, not task ground truth. It differs from $\tau^D$, the full dense run from the prompt, used for full-dense fidelity. The fixed-phase ranking tests use $\Udense$ as measured downstream effect; task repair is evaluated separately. If all compressed FFN events after $t_0$ are restored, $\tau^{R,A}=\tau^{D\mid C}$; frozen regression tests verify this implementation invariant.

\begin{algorithm}[H]
\caption{\textbf{Same-State Causal FFN Restoration.}}
\label{alg:ssr}
\small
\begin{algorithmic}[1]
\Require prompt $p$; FFNs $F^D_\ell,F^C_{\ell,t}$; frozen events $A=W\times S$; fixed sampler
\Ensure compressed continuation $\tau^C_p$ and restored continuation $\tau^{R,A}_p$
\State run compressed inference to $t_0=\min W$; save the complete snapshot $s^C_{p,t_0}$
\State continue $s^C_{p,t_0}$ with compressed FFNs to obtain $\tau^C_p$
\State clone the same snapshot: $s^R_{p,t_0}\gets s^C_{p,t_0}$
\For{$t=t_0,\ldots,T$}
    \For{each FFN layer $\ell$ in model order}
        \State compute the current restoration-branch FFN input $h^R_{t,\ell}$
        \If{$(t,\ell)\in A$}
            \State $z\gets F^D_\ell(h^R_{t,\ell})$ \Comment{restore the original FFN}
        \Else
            \State $z\gets F^C_{\ell,t}(h^R_{t,\ell})$
        \EndIf
        \State continue the restoration branch with $z$ and unchanged non-FFN computation
    \EndFor
\EndFor
\State \Return $\tau^C_p,\tau^{R,A}_p$
\Statex \textbf{Invariant:} same snapshot at $t_0$; same sampler and non-FFN computation; frozen $A$; no dense hidden state is copied.
\end{algorithmic}
\end{algorithm}
\vspace{-7.07pt}

\subsection{What local error magnitude misses: state and correction structure}
Local error magnitude can be useful while still discarding two distinct kinds of downstream-relevant information: where the compressed model currently is, and which feature-space correction is missing. We test these possibilities separately. The state test is controlled predictive evidence after phase is fixed; the correction-structure test is an intervention at fixed norm.

\paragraph{State beyond phase.}
We fix denoising phase, mask count where applicable, restored layers, and horizon, then ask whether compressed-side state observables still rank $\Udense$. Observables are computed entirely from the compressed branch and include progress/mask statistics, release confidence, release margin, entropy, active-token summaries, and prompt length; dataset identity and gold answers are excluded. The release margin is the top-1 minus top-2 logit gap at the token selected for release; we use its negative so larger values mean lower local release certainty. This is controlled predictive evidence that compressed-side state contains ranking information beyond fixed phase, not a causal manipulation of the observable.

\paragraph{Same norm, altered structure.}
Original-FFN restoration adds $\delta$ to the compressed FFN output. We compare it with norm-matched $P_{\ell,t}\delta$ and, where used, $-\delta$ over the same events. $P_{\ell,t}$ is a frozen cyclic feature permutation; because it is orthogonal, it preserves $\|\delta\|_2$ exactly while changing feature-space direction/alignment. Different downstream outcomes at equal norm establish that correction magnitude alone is insufficient; they do not identify a complete structural causal model. Appendix~\ref{app:mechanism} gives the exact control construction.

\subsection{Close the loop: freeze and transfer a restoration window}
We use source measurements to choose one global restoration window, freeze it, and then evaluate held-out prompts without search or refitting. Algorithm~\ref{alg:ssr} is the operator-level intervention; Algorithm~\ref{alg:freeze} is the separate calibration and transfer protocol. For a start $t$ and horizon $H$, define the temporal window and joint intervention set
\[
W(t,H)=\{t,\ldots,t+H-1\},
\qquad
A(t,H)=W(t,H)\times S .
\]
For a window $W$, $E_p(W,S)$ denotes Eq.~\ref{eq:nmse} after summing numerator and denominator over $t\in W$ before taking the ratio. Thus Algorithm~\ref{alg:ssr} defines the intervention, while Algorithm~\ref{alg:freeze} chooses and freezes its one global window.

\begin{table}[H]
\caption{\textbf{Frozen evaluation regimes.} $t/H$ denotes start transition / horizon. The last column reports estimated whole-model MAC saving relative to dense, from compressed-only to \SSR.}
\label{tab:regimes}
\centering\small
\begin{tabular}{@{}lcccc@{}}
\toprule
\textbf{Family / regime} & \textbf{Compressed FFNs} & \textbf{\SSR $t/H$} & \textbf{\NMSE $t/H$} & \textbf{Comp.$\rightarrow$\SSR saving}\\
\midrule
\LLaDA c24 & 24/32 & 27/4 & 28/4 & 42.3$\rightarrow$36.8\%\\
\LLaDA c14 & 14/32 & 27/4 & 28/4 & 24.7$\rightarrow$21.5\%\\
\Dream c21 & 21/28 & 17/2 & 30/2 & 42.1$\rightarrow$39.3\%\\
\bottomrule
\end{tabular}
\end{table}

\vspace{-5.6pt}
\begin{algorithm}[H]
\caption{\textbf{Freeze one global restoration window, then transfer.}}
\label{alg:freeze}
\small
\begin{algorithmic}[1]
\Require fixed compressed layers $S$; starts $\mathcal{T}$; broad horizon $H_0$; horizon set $\mathcal{H}$
\Require disjoint source sets $D_{\rm scr},D_{\rm conf},D_{\rm hor}$; frozen local-error set $D_{\rm NMSE}$; held-out $D_{\rm eval}$
\Ensure frozen $W_{\SSR},W_{\NMSE}$ and held-out outputs
\Statex \textbf{Source calibration (task correctness unused)}
\For{$t\in\mathcal{T}$}
    \State $a_t\gets \operatorname{Agg}_{p\in D_{\rm scr}}\Udense\!\left(p,W(t,H_0)\times S\right)$
\EndFor
\State $B\gets$ prespecified top starts by $a_t$
\State $t^\star\gets \arg\max_{t\in B}\operatorname{Agg}_{p\in D_{\rm conf}}\Udense\!\left(p,W(t,H_0)\times S\right)$
\State $H^\star\gets$ frozen horizon rule on $\{\Udense(p,W(t^\star,H)\times S):p\in D_{\rm hor},H\in\mathcal{H}\}$
\State $W_{\SSR}\gets W(t^\star,H^\star)$
\State $W_{\NMSE}\gets\arg\max_{W:\,|W|=H^\star}\operatorname{Agg}_{p\in D_{\rm NMSE}}E_p(W,S)$
\State \textbf{freeze} $W_{\SSR}$ and $W_{\NMSE}$
\Statex \textbf{Frozen transfer}
\For{$p\in D_{\rm eval}$}
    \State $Y_{\SSR}\gets \operatorname{SSR}(p,W_{\SSR}\times S)$ \Comment{Algorithm~\ref{alg:ssr}}
    \State run compressed and $W_{\NMSE}$ comparators; score prespecified endpoints
\EndFor
\Statex \textbf{Boundary:} no task correctness before the freeze; no per-prompt search/refit after it.
\end{algorithmic}
\end{algorithm}

\paragraph{Window selection.}
Candidate sets, split roles, and aggregation rules are frozen before the corresponding source-measurement outcomes. For \LLaDA, the broad screen uses $H_0=4$, confirms only the prespecified top two starts, then chooses $H^\star\in\{1,2,4\}$ on a third source set; exact candidates are in Appendix~\ref{app:setup}. The \SSR objective is an equal-dataset macro of 10\%-winsorized $\Udense$ effects. The local-\NMSE comparator applies the same frozen equal-dataset aggregation to $E_p(W,S)$ and selects the highest-scoring length-$H^\star$ window. Calibration uses same-snapshot dense continuations but never task correctness, and MMLU remains unopened until windows and predictors are frozen.

\paragraph{Frozen execution.}
After $W_{\SSR}$ is frozen, Eq.~\ref{eq:restoreop} defines a single mixed-operator trajectory from the prompt: compressed FFNs execute outside the frozen window and original FFNs execute at treated events. No same-snapshot dense continuation or compressed counterfactual branch is required to produce that output. Original FFN weights remain resident. We therefore study selective operator execution, not parameter compression or measured latency speedup.

For \LLaDA, the local-\NMSE-selected $t=28,H=4$ schedule is also the literal final-four-transition schedule $\{28,29,30,31\}$. Thus the equal-horizon comparator simultaneously tests the simple heuristic of restoring only the final four denoising transitions.

\paragraph{Evaluation endpoints.}
Full-dense fidelity and task repair are separate. A frozen multiple-choice parser returns one admissible label or invalid; correctness requires the gold label, while validity is reported separately. A repair opportunity is full-dense correct/compressed incorrect, and conditional repair is the fraction repaired by \SSR; positive $\Udense$ need not cross this task boundary.

\section{Experiments}
\subsection{Setup and frozen protocol}

\paragraph{Models and compression testbed.}
We evaluate frozen \LLaDA-8B-Instruct and \Dream-v0-Instruct-7B revisions \citep{nie2025llada,ye2025dream} with deterministic 32-transition, length-32 generation at temperature 0. Both use frozen instances of one structured FFN-width testbed (Table~\ref{tab:regimes}). \LLaDA uses nested 1,024/1,536-channel slices from native width 12,288 with a transition-zero static anchor; \Dream uses 4,608/6,912-channel slices from native width 18,944 with an outcome-blind activation-aware ordering and shifted predictor-row routing. Exact revisions, layer sets, channel construction, anchors, and schedules are in Appendix~\ref{app:setup}. We hold the compression construction fixed to isolate the causal question rather than compare compressors; the cross-model axis tests the same estimand in two independently developed DLM families. c24/c21 are aggressive 75\%-FFN stress tests; c14 is a separately frozen moderate boundary/evidence regime.

\paragraph{Source and held-out populations.}
Source calibration uses ARC-Challenge, ARC-Easy, SciQ, PIQA, and HellaSwag with disjoint screen/confirmation/horizon cohorts (8/8/4 prompts per dataset). c24 is the aggressive \LLaDA stress test, c14 probes the moderate boundary and fixed-phase evidence, and c21 tests transfer to an independently developed DLM family. The matched final population has 297 prompts: 48 per source dataset plus 57 MMLU subject-level prompts, with MMLU unopened during selection. \LLaDA's initial 594-prompt manifest was reduced outcome-blind to 297 before final-condition inference; frozen policies were unchanged (Appendix~\ref{app:provenance}).

\paragraph{Statistics and compute scope.}
The prompt is the inferential unit. We use paired prompt bootstraps, exact discordant-pair tests, Wilson intervals, permutation tests, and prompt-grouped cross-validation (Appendix~\ref{app:stats}). Savings are analytical whole-model direct-dispatch MAC estimates, not measured latency, memory, or parameter storage; anchor construction and frozen restoration are included, offline calibration is not.

\subsection{Evidence beyond local error magnitude}

{\setlength{\intextsep}{4pt}
\begin{figure}[H]
    \centering
    \includegraphics[width=\linewidth]{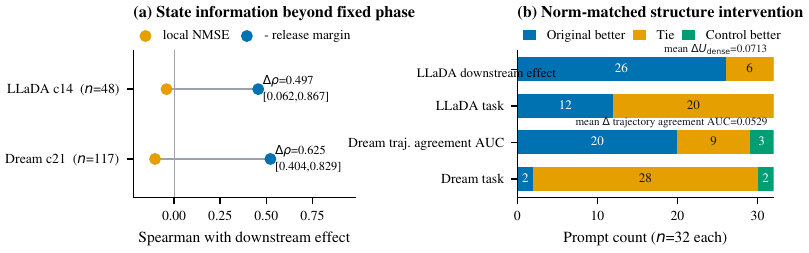}
    \caption{\textbf{Local error magnitude omits state and correction-structure information.} Left: at fixed phase, negative release margin ranks downstream effect more strongly than local \NMSE. Right: exact restoration is compared with a norm-matched structure control. State evidence is predictive; the structure test is interventional. Details are in Appendix~\ref{app:mechanism}.}
    \label{fig:mechanism}
\end{figure}}

\paragraph{State beyond phase.}
The finding is not that DLM state is generally informative; prior adaptive-computation work already establishes that. Rather, after phase is held fixed, a compressed-side state observable ranks the measured downstream effect of missing FFN computation while local FFN \NMSE is nearly uninformative. With five remaining masks and $H=4$, negative release margin gives \LLaDA $\rho=.456$ vs. $-.041$ for local \NMSE and \Dream $.522$ vs. $-.103$ (Figure~\ref{fig:mechanism}). Negative release margin emerged during earlier exploratory development and was prospectively frozen for the reported confirmatory tests. A fresh \LLaDA ARC-Challenge confirmation fixed $t=27$, $H=4$, five masks, the layer set, and this scalar before outcomes ($n=32$), giving $\rho=.500$, one-sided permutation $p=.0022$, 95\% CI [.190,.743]. This is controlled predictive evidence, not a causal intervention on release margin.

\paragraph{Correction structure beyond norm.}
Norm-matched controls change correction structure while preserving its norm and treated events. Original-FFN restoration wins 26/32 prompts with zero reversals on the \LLaDA same-snapshot downstream-effect endpoint, while \Dream trajectory specificity transfers with 20/32 original-restoration wins against 3 reversals; \Dream task specificity is null (2/28/2). Thus correction magnitude alone is insufficient, although these controls do not identify a complete structural causal model.

Together, the tests show that local FFN error magnitude discards two downstream-relevant sources of information: where the model currently is and which computation is missing. Predictor generalization remains weak (negative leave-dataset-out $R^2$), so we make no router claim; aggregation sensitivities preserve the state$>$\NMSE ordering, while a broad late-versus-early CI crosses zero (Appendix~\ref{app:mechanism}).

\subsection{Frozen closed-loop outcomes: full-dense fidelity and task repair separate}

{\setlength{\intextsep}{4pt}
\begin{figure}[H]
    \centering
    \includegraphics[width=\linewidth]{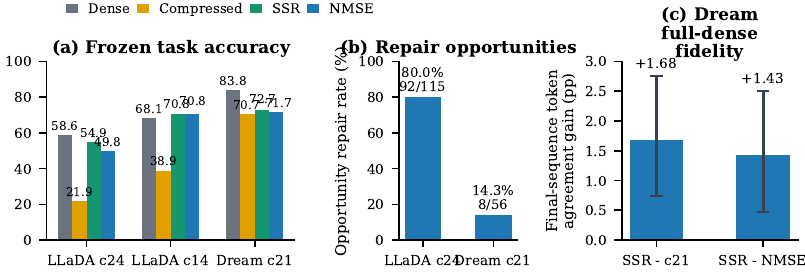}
    \caption{\textbf{Closed-loop outcomes separate task repair from full-dense fidelity.} (a) Frozen task accuracy. (b) Repair among dense-correct/compressed-wrong opportunities. (c) \Dream token-agreement gains with paired 95\% CIs. Aggressive c24 recovers 89.9\% of the dense--compressed accuracy gap while retaining 36.8\% estimated whole-model MAC saving; c14 is a task-equivalent boundary, and \Dream improves fidelity without a reliable task effect.}
    \label{fig:closedloop}
\end{figure}}
\vspace{-0.96pt}

{\setlength{\intextsep}{5pt}
\begin{table}[H]
\caption{\textbf{Closed-loop task accuracy under frozen equal-horizon schedules.} Accuracies and effects are pooled over prompts; effects are percentage points. The prespecified equal-dataset macro c24 effect is in the text; MAC savings are in Table~\ref{tab:regimes}.}
\label{tab:closedloop}
\centering\small
\setlength{\tabcolsep}{4pt}
\begin{tabular}{@{}lrrrrrr@{}}
\toprule
\textbf{Regime ($n$)} & \textbf{Dense} & \textbf{Comp.} & \textbf{\SSR} & \textbf{\NMSE} &
\textbf{\SSR$-$Comp. [95\% CI]} & \textbf{\SSR$-$\NMSE [95\% CI]}\\
\midrule
\LLaDA c24 (297) & 58.6 & 21.9 & 54.9 & 49.8 & +33.0 [27.6,38.4] & +5.05 [2.36,8.08]\\
\LLaDA c14 (72) & 68.1 & 38.9 & 70.8 & 70.8 & +31.9 [22.2,43.1] & 0.00\\
\Dream c21 (297) & 83.8 & 70.7 & 72.7 & 71.7 & +2.02 [$-.34$,4.38] & +1.01 [$-2.02$,4.04]\\
\bottomrule
\end{tabular}
\end{table}}
\vspace{-1.0pt}

\noindent\textbf{Aggressive \LLaDA (c24).}
\SSR recovers 89.9\% of the dense--compressed accuracy gap (prespecified equal-dataset macro +33.2\pp, 95\% CI [27.8,38.5]) while retaining an estimated 36.8\% whole-model MAC saving. It reaches 54.9\% accuracy versus 21.9\% compressed and 49.8\% for the equal-horizon local-\NMSE/final-four schedule, a +5.05\pp advantage (18 \SSR-only vs.\ 3 comparator-only correctness wins; exact two-sided $p=.00149$). It repairs 92/115 dense-correct/compressed-wrong opportunities. The recovery is predominantly validity-mediated: 98/99 \SSR-only correctness gains are invalid$\rightarrow$correct. Under severe FFN compression, the observed functional validity collapse is therefore highly reversible by one frozen four-transition window. We do not infer generic reasoning improvement from this experiment.

\par\smallskip\noindent\textbf{Moderate \LLaDA (c14).}
This regime defines an informative boundary: \SSR and matched local-\NMSE have identical prompt-level correctness at 70.8\% on all 72 prompts, showing that local approximation error can be sufficient for task selection in a less severe regime. Their parsed labels differ once where both are wrong; partial restoration above dense accuracy is not evidence of superiority over the original model.

\par\smallskip\noindent\textbf{\Dream (c21).}
\Dream isolates the distinction between restoration and task repair: task effect +2.02\pp (95\% CI [$-.34$,4.38]) is inconclusive, while full-dense final-sequence token agreement improves +1.68\pp (95\% CI [.74,2.75]) and trajectory agreement AUC +.0082. At near-matched MAC, \SSR exceeds \NMSE by +1.43\pp token agreement and +.0080 AUC. Conditional repair is 14.3\% (8/56), versus 80.0\% (92/115) in c24; a separately sampled frozen \Dream population likewise gives an exact task tie (6 vs.\ 6 discordances) but +1.20\pp token agreement (95\% CI [.18,2.25]). Thus restoration can repair a task boundary, tie a local-error selector, or improve dense fidelity without reliably changing the final decision.

\vspace{0.59pt}
\subsection{Claim-critical controls and robustness}
\vspace{-6.63pt}
\begin{table}[H]
\caption{\textbf{Claim-critical controls and robustness.} Rows are ordered by the simplest reviewer alternatives; Appendix~\ref{app:ablations} gives additional sensitivities.}
\label{tab:ablations}
\centering
\small
\renewcommand{\arraystretch}{0.93}
\begin{tabularx}{\linewidth}{@{}>{\raggedright\arraybackslash}p{0.23\linewidth}>{\raggedright\arraybackslash}p{0.32\linewidth}Y@{}}
\toprule
\textbf{Question} & \textbf{Controlled comparison} & \textbf{Result}\\
\midrule
Does phase/local error explain ranking? & fixed phase, masks, layers, horizon; state vs. local \NMSE & $\rho$: \LLaDA .456 vs. $-.041$; \Dream .522 vs. $-.103$\\
Would any same-norm correction work? & exact vs. norm-matched cyclic permutation; sign flip in App.~\ref{app:mechanism} & \LLaDA $\Udense$ 26/6/0; \Dream trajectory AUC 20/9/3; task 2/28/2\\
Is c24 just a late-window / more-compute effect? & \SSR 27--30 vs. \NMSE/final-four 28--31; equal $H=4$, identical est. MAC & \SSR +5.05\pp (18--3, $p=.00149$); all-40 \NMSE rescore keeps 28/4\\
Is one transition enough? & ARC-only, fixed $t=27$; $H=1$ vs. 4 & 1/12 vs. 7/12 repairs; six $H=4$-only discordances, $p=.03125$\\
Could parsing create the \Dream result? & parser-independent sequence/trajectory endpoints & token agreement +1.68\pp [.74,2.75]; trajectory AUC +.0082\\
Could selection leak held-out labels? & task-label-free objective; disjoint splits; MMLU unopened & frozen before held-out task evaluation; no per-prompt search/refit\\
\bottomrule
\end{tabularx}
\end{table}
\vspace{-8.04pt}
The controls reject the main simple alternatives: state ranking is not explained by fixed phase/local \NMSE; equal-norm perturbations rule out magnitude alone; c24 beats the literal final-four window at identical estimated MAC; $H=1$ does not reproduce the ARC pattern; \Dream fidelity is parser-independent; and selection is frozen before held-out evaluation. The all-40 rescore does not equalize calibration budgets, and c14 remains the task-superiority boundary.

\vspace{0.58pt}
\section{Discussion, Limitations, and Conclusion}
\vspace{0.59pt}
\noindent\textbf{Scientific implication.}
\SSR makes the downstream consequence of missing FFN computation measurable without importing a dense hidden trajectory. Across two DLM families, local magnitude alone is incomplete: compressed-side state adds ranking information after phase is fixed, and norm-preserving interventions show that correction structure matters. Closed-loop results further separate movement toward dense behavior from final task repair; aggressive \LLaDA recovers most of a severe compression loss while preserving substantial estimated compute savings, whereas \Dream improves dense fidelity without a conclusive task gain.

\vspace{0.63pt}
\par\smallskip\noindent\textbf{Implication for compression objectives.}
Local \NMSE measures immediate same-input mismatch, but iterative dynamics can absorb or amplify that mismatch differently across states. \SSR therefore provides a diagnostic target for testing downstream-aware compression proxies and compute-allocation rules: the useful target is the consequence of the missing computation, not reconstruction magnitude alone. The current calibration still uses the original FFNs and same-snapshot dense continuations, so this diagnostic should not be confused with a dense-free deployment selector.

\vspace{0.66pt}
\par\smallskip\noindent\textbf{Boundaries.}
c24 is predominantly validity-mediated recovery, c14 is an exact \SSR--\NMSE correctness tie, and the \Dream task effect is inconclusive. We intentionally hold one structured FFN-width construction fixed to isolate the intervention; cross-compressor generalization remains open. The evaluation covers two DLM families, deterministic 32-transition length-32 generation, and mainly multiple-choice endpoints. Same-snapshot dense continuations are calibration references rather than task ground truth; original FFNs remain resident and savings are analytical MAC estimates, not latency, memory, or parameter compression. Long-form generation and dense-free selection also remain open.

\vspace{0.66pt}
\par\smallskip\noindent\textbf{Conclusion.}
Local FFN approximation error is not the downstream consequence of removing that computation from an iterative DLM. \SSR isolates the latter on the current branch and shows that state/context and correction structure carry information that local magnitude omits. DLM compression should distinguish local mismatch, downstream consequence, and task repair rather than treating them as interchangeable objectives.

\clearpage
\section*{Reproducibility Statement}
Appendix~\ref{app:setup} records model revisions, the structured FFN-width construction, frozen compressed-layer sets, deterministic generation, source/final population construction, and window candidates. Appendix~\ref{app:stats} specifies endpoint and statistical conventions; Appendix~\ref{app:mechanism} gives state and correction-control construction; Appendix~\ref{app:ablations} gives claim-critical sensitivities; Appendix~\ref{app:compute} gives MAC accounting; and Appendix~\ref{app:provenance} maps artifact terminology to the paper. The released project artifacts include frozen manifests, raw predictions, and recomputation notebooks.

\section*{AI Use}
Generative AI tools were used to assist with literature search, research brainstorming and hypothesis refinement, feedback on experimental design and methodology, implementation and debugging of experimental code, interpretation of experimental results, scientific figure and manuscript development, drafting and language editing, and manuscript formatting. All experiments were executed by the authors, and AI-assisted code, analyses, citations, numerical results, scientific claims, and final manuscript content were checked against the underlying project artifacts and source literature. The authors take responsibility for the final content of this work.

\begingroup
\setlength{\bibsep}{4.29pt}
\renewcommand{\bibfont}{\fontsize{8.783pt}{10.4pt}\selectfont}
\makeatletter
\renewcommand\NAT@bibsetup[1]{\setlength{\leftmargin}{\bibhang}\setlength{\itemindent}{-\leftmargin}\setlength{\itemsep}{\bibsep}\setlength{\parsep}{0pt}\setlength{\topsep}{4.64pt}}
\makeatother

\input{references_manual}
\endgroup
\appendix
\input{appendix}

\end{document}

%% file: appendix.tex
{\makeatletter
\def\section{\@startsection {section}{1}{\z@}{-4.49ex plus -0.5ex minus -.2ex}{1.5ex plus 0.3ex minus0.2ex}{\large\sc\raggedright}}
\section{Setup, frozen protocol, and endpoint definitions}
\makeatother}
\label{app:setup}
\subsection{Models, compression, and frozen schedules}
\textbf{\LLaDA.} We use \texttt{GSAI-ML/LLaDA-8B-Instruct}, revision \texttt{08b83a6feb34df1a}\\
\texttt{6011b80c3c00c7563e963b07}. The model has 32 layers, hidden width 4096, and native FFN width 12,288. The aggressive c24 compressed layer set is
\[
\{0,2,3,4,6,7,8,10,11,12,14,15,16,17,19,20,21,23,24,25,27,28,29,31\}.
\]
Frozen nested channel orderings define narrow/wide widths 1,024/1,536. The 1,024-channel slice is a subset of the 1,536-channel slice at every compressed layer. Prompt and already released rows use the narrow slice; currently masked generated rows use the wider slice. At transition 0, the original FFN is evaluated densely to construct a row-wise dense-minus-sliced residual for each width; this residual is cached as a static anchor and its construction cost is included in the whole-model MAC ledger. The channel ordering is fixed before the \SSR experiments and is never selected from \SSR or task outcomes. Primary \SSR starts at transition 27 with horizon 4; the equal-horizon local-\NMSE/final-four schedule starts at 28. Moderate c14 is a separately frozen 14-layer subset of c24 and reuses the frozen 27/4 and 28/4 schedules.

\par\vskip2.60pt\relax
\noindent\textbf{\Dream.} We use \texttt{Dream-org/Dream-v0-Instruct-7B}, revision \texttt{9c73b03fe0c0fb2a}\\
\texttt{0ad2553e43c7e3ee8daa0170}. The model has 28 layers, hidden width 3584, and native FFN width 18,944. The aggressive c21 compressed layer set is
\[
\{0,2,3,4,6,7,8,10,11,12,14,15,16,18,19,20,22,23,24,26,27\}.
\]
Narrow/wide widths are 4,608/6,912. \Dream uses a model-native, outcome-blind activation-aware channel ordering: on source-only calibration trajectories, squared SwiGLU intermediate activation is accumulated over predeclared mask states and balanced roles, then combined with the corresponding down-projection column norm to produce one nested per-layer ranking. The wider slice is routed to predecessor positions associated with currently masked targets; other rows use the narrow slice. Transition-zero dense-minus-sliced anchors are constructed analogously and included in the MAC ledger. Primary \SSR starts at transition 17 with horizon 2; the local-\NMSE schedule starts at 30 with horizon 2.

Both families use deterministic 32-transition generation at temperature 0 with length 32. Source calibration uses ARC-Challenge, ARC-Easy, SciQ, PIQA, and HellaSwag. Screen, confirmation, and horizon cohorts are disjoint. Task correctness is unused during selection, and MMLU remains unopened until the windows and source-side predictors are frozen.

{\makeatletter
\def\subsection{\@startsection{subsection}{2}{\z@}{-2.13ex plus -0.5ex minus -.2ex}{0.8ex plus .2ex}{\normalsize\sc\raggedright}}
\enlargethispage{4pt}\subsection{Statistics and endpoints}
\makeatother}
\label{app:stats}
The prompt is the inferential unit. Accuracy and fidelity differences use paired prompt bootstrap intervals unless otherwise stated. Exact paired tests operate on discordant correctness indicators. Wilson intervals are used for proportions where noted; permutation tests are used for Spearman significance; predictor analyses use prompt-grouped cross-validation.

Final-sequence token agreement is mean token equality over all generated output positions relative to the full dense run $\tau^D$. Trajectory agreement AUC averages per-transition agreement with the corresponding full-dense trajectory over the frozen evaluation horizon. The released-aware variant evaluates agreement on positions already released by the sampler at each transition. These are graded full-dense fidelity endpoints, distinct from final multiple-choice correctness.

\clearpage
\addtolength{\topskip}{2.26pt}
\section{Held-out closed-loop results}
\vspace{-1.67pt}
\begin{table}[h]
\caption{\Dream \SSR versus \NMSE at essentially matched whole-model MAC cost on 297 prompts.}
\centering\small
\begin{tabular}{@{}lrr@{}}
\toprule
\textbf{Endpoint} & \textbf{Effect} & \textbf{95\% paired CI}\\
\midrule
Task correctness & .0101 & [$-.0202$,.0404]\\
Validity & .0236 & [.0000,.0505]\\
Final-sequence token agreement to full dense run & .0143 & [.0046,.0249]\\
Trajectory agreement AUC to full dense run & .0080 & [.0052,.0109]\\
Released-aware trajectory agreement AUC & .0095 & [.0064,.0129]\\
\bottomrule
\end{tabular}
\end{table}

\vspace{-2.37pt}
\begin{table}[h]
\caption{\LLaDA c24 \SSR versus equal-horizon \NMSE on the frozen 297-prompt population.}
\centering\small
\begin{tabularx}{\linewidth}{@{}lrY@{}}
\toprule
\textbf{Endpoint} & \textbf{\SSR$-$\NMSE} & \textbf{Paired evidence}\\
\midrule
Task accuracy & +5.05\pp & 18 vs.\ 3 discordances; exact $p=.00149$; bootstrap CI [2.36,8.08]\pp\\
Whole-model MAC saving & 0.00\pp & both 36.76\% under the common ledger\\
\bottomrule
\end{tabularx}
\end{table}

\vspace{-1.79pt}
\section{Mechanistic evidence and controls}
\label{app:mechanism}
\subsection{State ranking}
The frozen \LLaDA primary statistic is an equal-dataset macro across three 16-prompt strata; the frozen \Dream primary statistic is prompt-pooled over 117 prompts. Under one common prompt-pooled rule, the state$>$\NMSE ordering is preserved. Under one common equal-dataset macro rule, \LLaDA is .456 vs.\ $-.041$ and \Dream is .506 vs.\ $-.185$. We therefore claim cross-family directional consistency, not equality of absolute correlations.

\vspace{10.99pt}
A fresh \LLaDA ARC-Challenge confirmation fixed transition 27, horizon 4, five remaining masks, the compressed layer set, and negative release margin before outcomes. On $n=32$ prompts, Spearman $\rho=.500$, one-sided permutation $p=.0022$, 95\% CI [.190,.743].

Prompt-grouped held-out predictors support ranking but not calibrated deployment. Negative or weak leave-dataset-out $R^2$ motivates the main paper's ranking-only language and rules out a learned-router claim.

\subsection{Norm-matched correction structure}
At each treated event, the exact restoration correction is $\delta$. The cyclic-permutation control applies a frozen feature permutation $P_{\ell,t}$ to $\delta$ at the same events. Because $P_{\ell,t}$ is orthogonal, $\|P_{\ell,t}\delta\|_2=\|\delta\|_2$ exactly; sign-flip controls use $-\delta$. Thus correction norm and treated locations are held fixed while feature-space direction/alignment changes.

\vspace{-1.93pt}
\begin{table}[h]
\caption{Correction-structure specificity. Counts are original-FFN restoration better / tie / structure-control better.}
\centering\small
\begin{tabular}{@{}lrr@{}}
\toprule
\textbf{Family / endpoint} & \textbf{Counts} & \textbf{Mean exact$-$control}\\
\midrule
\LLaDA same-snapshot downstream effect & 26/6/0 & .0713\\
\LLaDA task correctness & 12/20/0 & .3750 indicator\\
\Dream trajectory agreement AUC & 20/9/3 & .0529\\
\Dream final-sequence token agreement & 5/23/4 & .0410\\
\Dream task correctness & 2/28/2 & .0000\\
\bottomrule
\end{tabular}
\end{table}

\clearpage
\addtolength{\topskip}{0.46pt}
\section{Claim-critical ablations and sensitivities}
\label{app:ablations}
\vspace{-2.36pt}
\begin{table}[H]
\caption{Horizon and placement controls.}
\centering\small
\begin{tabularx}{\linewidth}{@{}p{0.30\linewidth}p{0.30\linewidth}Y@{}}
\toprule
\textbf{Comparison} & \textbf{Result} & \textbf{Interpretation}\\
\midrule
\LLaDA $H=4$ vs.\ $H=1$, ARC-sensitive failures ($n=12$) & 7 vs.\ 1 rescues; 6 $H=4$-only discordances; exact $p=.03125$ & one restored transition is insufficient on this ARC cohort\\
Exact vs.\ permuted correction & positive exact-control downstream effect & extra correction norm alone does not explain the effect\\
Broad late$-$early \LLaDA macro & positive point estimate; CI crosses zero & do not universalize a late-transition preference\\
\Dream task specificity & 2/28/2 & correction structure need not change final task correctness\\
\bottomrule
\end{tabularx}
\end{table}

\vspace{-0.85pt}\paragraph{\NMSE-selector data-budget sensitivity.}
The executed \LLaDA \NMSE baseline selected its start from a frozen 6-prompt-per-dataset subset of the 8-prompt broad screen, with the \SSR-selected horizon fixed at four transitions. Recomputing the identical energy-weighted selector on all 40 broad-screen prompts leaves the winner unchanged: $t=28,H=4$ has macro \NMSE .12107 versus .11693 for $t=27,H=4$ and .11360 for $t=26,H=4$. A separate post-hoc sensitivity using stored per-prompt \NMSE values on the disjoint 40-prompt confirmation cohort also ranks $t=28$ above $t=27$ (.11457 vs.\ .10894). The latter cannot reproduce the exact energy-weighted aggregation because the necessary per-prompt SSE/energy components were not archived there. These checks show that the chosen start is not fragile to the smaller selector subset; they do not establish a fully equal calibration budget between \SSR and \NMSE.

\vspace{-0.45pt}
\section{Failure types and compute accounting}
\label{app:compute}
The ledger includes attention projections and score/value products, FFNs, LM head, transition-zero anchor construction, and frozen restoration. It estimates arithmetic for a functionally equivalent

\vspace{-7.52pt}
\begin{table}[h]
\caption{Compressed$\rightarrow$\SSR outcome transitions on matched populations.}
\centering\small
\begin{tabular}{@{}lrr@{}}
\toprule
\textbf{Transition} & \textbf{\LLaDA c24} & \textbf{\Dream c21}\\
\midrule
Invalid$\rightarrow$correct & 98 & 7\\
Valid-wrong$\rightarrow$correct & 1 & 2\\
Correct$\rightarrow$wrong & 0 & 2\\
Correct$\rightarrow$invalid & 1 & 1\\
Invalid$\rightarrow$valid-wrong & 26 & 1\\
Correct$\rightarrow$correct & 64 & 207\\
Other unchanged wrong/invalid & 107 & 77\\
\bottomrule
\end{tabular}
\end{table}

\vspace{-5.94pt}
\begin{table}[h]
\caption{Estimated whole-model MAC saving under one common accounting convention.}
\centering\small
\begin{tabular}{@{}lr@{}}
\toprule
\textbf{Family / condition} & \textbf{Mean saving}\\
\midrule
\LLaDA c24 compressed & 42.26\%\\
\LLaDA c24 \SSR & 36.76\%\\
\LLaDA c24 \NMSE & 36.76\%\\
\LLaDA c14 \SSR & 21.45\%\\
\Dream c21 compressed & 42.08\%\\
\Dream c21 \SSR & 39.29\%\\
\Dream c21 \NMSE & 39.24\%\\
\bottomrule
\end{tabular}
\end{table}

\vspace{9.14pt}
\noindent optimized direct-dispatch implementation. It is not a measurement of wall-clock latency, memory, parameter storage, or hardware FLOPs. Offline calibration is excluded from per-inference saving, and original FFN weights remain resident.

\clearpage
\addtolength{\topskip}{1.40pt}
\section{Artifact and reproducibility boundary}
\label{app:provenance}
Historical project artifacts use \texttt{CFR}, \texttt{global\_cfr}, and \texttt{CFRController} for the exact intervention denoted \SSR in this paper; this is a terminology change, not an algorithmic change. Artifacts also use \texttt{step} for what the paper calls a denoising transition. Archived columns named \texttt{final\_token\_agreement} are the quantity called final-sequence token agreement in the paper: mean token equality over all generated output positions, not agreement of only the last token.

The \LLaDA paper-mode notebook initially instantiated a 594-prompt final manifest (480 source-family prompts plus 114 MMLU records). Before final-condition inference, the 12-hour execution budget triggered an archived outcome-blind reduction to 297 using the already-defined stable-hash selector; no model outcomes were inspected and the compressor, sampler, parser, \SSR policy, and \NMSE policy were unchanged by that reduction. All headline \LLaDA statistics use the frozen 297-prompt population.

The frozen Stage59 and Stage59B method artifacts match, and Stage60 disables temporal, layer, channel, width, predictor, and test-set selection during cross-family evaluation. Matched Stage57/Stage60 populations are identity-audited. These checks support the source$\rightarrow$confirm$\rightarrow$horizon$\rightarrow$freeze chronology used in the main paper.